# Matriarchy Classics: A Computational Approach to Swap Only Gendered Words in Classic Novels

**Rachel Lea Draelos, MD, PhD**
Glass Box Medicine
rachel.draelos@glassboxmedicine.com

## Abstract

This paper proposes a trustworthy computational method to gender-swap public domain novels, with programmatic guarantees that only gendered words are inverted and all other words are left unchanged. This may seem like a simple task, but both large language models and naïve rule-based methods fail to complete it. The proposed method includes a custom whitespace-preserving tokenizer, gender-swap mappings across plural and possessive forms, pronoun handling, and a character renaming strategy. It shortens the time required to gender-swap a novel from weeks of manual labor to minutes. The method is demonstrated on two public-domain novels. *Pride and Prejudice* is transported to matriarchal Regency England, where inheritance flows through daughters, mothers rule the household, and a frantic Mr. Bennet is determined to secure advantageous marriages for his sons. *The Great Gatsby* pivots to a matriarchal Jazz Age where newly-rich Ms. Gatsbee throws lavish parties in the hopes of reuniting with her lost love, David, who is now married to the violent Thelma Buchanan. Gender-swapped novels could serve as useful tools to investigate the gender biases of readers, authors, societies, and AI systems.

## Introduction

Publishers have released dozens of gender-swapped versions of classic novels, from *Olivia Twist* by Lorie Langdon to *One for All* by Lillie Lainoff. However, critically, these are retellings—with every sentence, the author has chosen to what extent the new characters conform to or subvert gender stereotypes. And there may be more conformity than the gender-swap premise implies: a computational analysis of gender-bending fanfiction found that male characters turned female are often rewritten to conform with feminine stereotypes[1].

Original works of fiction have imagined matriarchal or gender-swapped societies for over a century, including *Sultana's Dream* (1905)[2], *Herland* (1915)[3], and *Egalia's Daughters* (1977)[4]. Gender-swapping classic novels has been reported by parents who dynamically perform the gender inversion while reading out loud to their children[5]. *Gender Swapped Fairy Tales* by Karrie Fransman and Jonathan Plackett[6] inverts only gendered language in classic fairy tales. Kim Knight's classroom gender bias exercise[7] involves using the digital Regender tool[8] created by Ping Yee to gender-swap web pages. Gender-swapped text has also been used to quantify and mitigate AI gender bias[9–11] and evaluate reader biases[12,13].

In the spirit of this prior work, the Matriarchy Classics project inverts only gendered words in classic novels, and leaves all other words unchanged. To the best of the author's knowledge, no prior work reports gender-only inversion of full-length novels, likely because carrying out this task by hand requires weeks of painstaking labor for each book. The primary contribution of this paper is a computational method that shortens the time required from weeks to minutes, with programmatic guarantees that only gendered words will be changed—a crucial guarantee for the efficacy of the project. The method is demonstrated on two public-domain novels, *Pride and Prejudice* and *The Great Gatsby,* with the gender-swapped versions made freely available as published ebooks through the project website, www.matriarchyclassics.com.

When the author first began this project six years ago, it seemed that developing a computational method to gender-swap novels would be trivial—merely a few lines of Python exchanging "he" for "she" and so on. Indeed, a few lines of Python *are* sufficient to produce a barely-readable gender-swapped atrocity, in which replacement of "man" with "woman" results in nonsense transformations like "manor" to "womanor." Producing a literary-quality gender swap for an entire novel while maintaining perfect grammar, preserving punctuation and formatting, ensuring appropriate character naming, and programmatically guaranteeing that *only* gendered words have been altered proves to be a more challenging problem, analogous to a specialized form of algorithmic translation—one that even modern large language models cannot complete successfully.

The proposed method may be useful for creating a gender-swapped canon of classic novels that could serve as tools in future gender bias research.

# Methods

The Matriarchy Classics software swaps only gendered words, leaving all other words unchanged.

## Ineffectual methods

Simply inserting the text of a classic novel into a large language model (LLM) like ChatGPT, Claude, or Gemini, and prompting the LLM to swap only gendered words does not work. The baked-in gender biases make it difficult for the model to reliably swap only gendered words, due to the statistical unlikelihood of certain resulting phrases. In order for the Matriarchy Classics reading experience to work, the reader must be able to trust that *only* gendered words have been changed, and nothing else. If the reader is constantly questioning whether a sentence reads weirdly because of revealed gender biases or because of an LLM-introduced error, the project experientially collapses.

As examples of how LLMs fail at this task, providing ChatGPT (7/14/2026) with the beginning of Chapter 1 of *The Great Gatsby* along with the prompt, "Swap only the gendered words and leave all other words unchanged" yields an interesting failure case. Consider this sentence from the original unchanged text: "I had a dog—at least I had him for a few days until he ran away—and an old Dodge and a Finnish woman, who made my bed and cooked breakfast and muttered Finnish wisdom to herself over the electric stove." ChatGPT gender-swaps the dog correctly, but it does *not* gender-swap the Finnish woman, because it apparently struggles to conceptualize a man cooking breakfast. Other experimentation with LLMs revealed instances where the models changed character names inconsistently or failed to swap clothing-related nouns culturally associated with gender.

There are two reasons that this gender inversion task is difficult for LLMs. First, precise gender inversion is a deterministic problem, but LLMs are inherently stochastic models that probabilistically predict the next word[14]. It is not ideal to use a stochastic model to solve a deterministic problem, as introduced errors will be random. Second, LLMs are themselves gender-biased because of extensive gender bias present in their training data, and no amount of post-training completely eliminates these biases[15,16]. Overall, using an LLM for this gender inversion task would mandate prohibitively onerous proofreading of the swapped text against the original to identify the random errors. A simple diffcheck would not aid proofreading, as it would mostly catch desired gender inversions, intermixed randomly with the errors.

In the remainder of the Methods section, I describe an alternative method, primarily rule-based, for trustworthy textual gender inversion that programmatically guarantees that *only* gendered words are swapped.

## Overview of the Matriarchy Classics method: computational steps

First the text is tokenized (split into its component words) using a custom punctuation and whitespace-preserving tokenizer. Next, all gendered words and character names are swapped. Each word in the book is only allowed to be changed once, to avoid looping changes that undo the gender-swap (e.g., “man” to “woman” getting re-transformed from “woman” back to “man”).

## Tokenization, whitespace, and punctuation

Consider the following text:

```
“I’d like to read a book,” said Alice.
```

The Python nltk tokenizer, a popular choice for tokenization, parses that sentence as:

```
['“', 'I', '’', 'd', 'like', 'to', 'read', 'a', 'book', ',', '”', 'said',
'Alice', '.']
```

The problem with this kind of tokenization is that it becomes difficult to reconstruct the punctuation and whitespace appropriately. Joining all of the pieces with a space produces a mess:

```
“ I ’ d like to read a book , ” said Alice .
```

The proposed method implements a custom tokenizer that preserves whitespace and punctuation. When applied to the example sentence it produces:

```
 ['“', 'I’d', ' ', 'like', ' ', 'to', ' ', 'read', ' ', 'a', ' ', 'book', ',”
', 'said', ' ', 'Alice', '.']
```

In this custom tokenization, every punctuation mark and whitespace character is included explicitly in the tokenizer’s output. The remainder of the code then iterates through the tokenizer’s output to change only gendered words, and afterwards the final sentence can be reconstructed by directly joining the pieces without spaces.

The software also includes a specific function to ensure that whitespace characters at the beginnings and ends of sentences are preserved, which preserves paragraph breaks.

## Gender-swapping words

### Gendered word mappings

The proposed method leverages gender-swap mappings for a variety of English words including animals (e.g., buck/doe, bull/cow, cock/hen), titles (e.g., king/queen, emperor/empress, headmaster/headmistress), familial relationships (e.g., mother/father, sister/brother, grandfather/grandmother), adjectives (e.g., bearded/haired, manly/womanly, beautiful/handsome), clothes (e.g., shirtwaist/shirt, petticoat/trousers), occupations (e.g., monk/nun, policeman/policewoman), and miscellaneous gendered words (e.g., god/goddess, hero/heroine, fraternity/sorority).

### Handling plural and possessive forms

The prototype version of the software leveraged a mapping that specified only the singular version of each noun (e.g., man/woman). Plural and possessive forms were created programmatically with the addition of s's and apostrophes. This immediately led to grammatical errors and nonsense words like "mans" instead of "men."

Because English has so many irregular plural and possessive forms, the final proposed method guarantees a grammatically correct gender swap through a curated mapping that includes every plural and possessive form of every target word, yielding over 930 mappings. Table 1 includes examples.

**Table 1.** Example mappings for handling irregular plural and possessive gendered words.

| Original | Swap To | Original | Swap To |
|---|---|---|---|
| man | woman | lad | lass |
| woman | man | lass | lad |
| men | women | lads | lasses |
| women | men | lasses | lads |
| man's | woman's | lad's | lass's |
| woman's | man's | lass's | lad's |
| men's | women's | lads' | lasses' |
| women's | men's | lasses' | lads' |

### Bidirectional vs unidirectional swaps

By default, all swaps are bidirectional (e.g., buck always becomes doe, and doe always becomes buck) but in some special cases, the swap is not bidirectional—e.g., "gown" becomes "coat" but "coat" does not become "gown." Similarly, "mayoress" is changed to "mayor" but if the word is already "mayor" this is not changed to "mayoress" because to a modern reader, it would be stranger to read about a woman who is a "mayoress" than a woman who is a "mayor."

### Imperfect swaps

Sometimes a word is gendered, but there is no exact opposite in English. For example, the mapping specifies that “chambermaid” should become “manservant,” and that “nymph” should become “satyr” although neither of those is a perfect inverse.

### Case-sensitive swapping for certain polysemous gendered words

By default all swaps are done regardless of the case of the original word. However, for a few select words a case-sensitive swap is performed instead: only uppercase “Count” is switched to “Countess” (to avoid “I count the apples” becoming “I countess the apples”), and only uppercase “Miss” becomes “Mister” (to avoid “I miss you terribly” becoming “I mister you terribly”).

### Case preservation

Books include all sorts of capitalization: lowercase, ALL CAPS, and Capitalized. The code ensures that the case is preserved throughout all swaps.

## Pronoun handling with language modeling

Nearly all pronoun inversions can be done using simple rules. There are two exceptions:

- “her” to “him” or “his”;
- “his” to “her” or “hers.”

For example, “I like her,” should become “I like him,” but “Her coat is blue,” should become “His coat is blue.”

Similarly, “His coat is blue” should become “Her coat is blue,” while “That coat is his” should become “That coat is hers.”

Correctly resolving these scenarios strictly requires language modeling, because the resolution is context-dependent. There are numerous options for prompting an LLM to perform the pronoun resolution. Prompting the LLM to edit the sentence directly often leads to poor performance and unpredictable errors, for the reasons previously discussed.

The proposed method first uses a rule to swap “her” to the token “him/his,” and another rule to swap “his” to the token “her/hers.” Then an LLM is prompted to output only the grammatically correct pronouns in a comma-separated list—effectively turning each relevant sentence into a grammar multiple-choice question. The pronouns the LLM lists out are then inserted into the sentence in the appropriate places using rules, to create the final sentence. A final rule-based step verifies that the pronoun swapping did indeed result in a swap, and that the outputted pronoun is indeed one of the allowed swap choices. If a mistake is found, the software writes it into an error log for manual review.

## Inbuilt programmatic guarantees

The rule-based software programmatically guarantees that (1) only words, character names, and pronouns appearing in the gender swap map will be swapped, and all other words will be unchanged; and (2) all punctuation and whitespace including paragraph breaks will be exactly preserved.

## Overview of the Matriarchy Classics method—human process

The previous sections described the computational method according to the order of operations that the computer performs when doing the automated gender-swap. However, the overarching process of producing a gender-swapped novel still requires human involvement:

(1) Run a script that automatically extracts proper nouns along with example sentences.
(2) Review the output of this step to determine which proper nouns are character names. There are vastly more character names in a novel than memorable characters. *Pride and Prejudice* has over fifty named characters and *Gatsby* has over eighty.
(3) Specify a gender-swap mapping for all gendered character names.
(4) Run the Matriarchy Classics software described above on the full text of the novel, to perform the gender swap.
(5) Review the gender-swapped text for subtleties that could benefit from manual editing. Examples of manual edits are discussed further in the Results.

The next subsections provide additional detail about character naming.

### Gender-inverted character names

First names in English often connote gender. Character names are critical for the reading experience of the gender-swapped text. It would be possible to ask an LLM to create gender-inverted character names without any human involvement, but this does not always yield the best reading experience.

The following three options for character renaming proved unsatisfactory:

1) Leaving the names alone, and letting the reader deal with reading about a "he" Elizabeth or a "she" Tom. This approach made it difficult to follow the narrative or keep track of the intended gender of the characters.
2) Changing the names to totally unrelated gender-swapped names, e.g., "Jane" to "William." This created a different kind of tension, namely a nagging desire to constantly cross-reference the original text to double check which original character was being referenced in a particular paragraph.
3) Swapping the names of famous "paired" opposite-gender characters in the original book, e.g., Elizabeth becomes Darcy and Darcy becomes Elizabeth. Surprisingly, this also made it difficult to read the swapped text, because of the disconnect between preexisting mental images of the characters and their new roles in the swapped story.

The approach ultimately settled on was to change character names to phonetically-related, period-plausible, opposite-gender names (e.g., Kitty to Kit, Louisa to Lewis, Daisy to David, Jay to Jane). If no satisfactory phonetically-related name was available, then a non-phonetically-related, period-appropriate, opposite-gender name was chosen instead, ideally with some other conceptual relation to the original name. For example, there was not a great male option for "Elizabeth"—something like "Eleazor" sounds too obscure and Biblical. Elizabeth is further referred to as Eliza/Lizzie/Lizzy throughout. Thus, a period-plausible male name with an analogous range of nicknames was ultimately selected: Theodore/Theo/Teddie/Teddy.

Although Darcy can be a woman's name, the character of Mr. Darcy is so iconic that it was necessary to change the name to something else to better imagine her as a woman. Although Darcy is a surname, the first name "Daphne" was ultimately chosen as a replacement because use of that name made it recurringly clear that the character is a woman, while retaining enough phonetic similarity to "Darcy" to be identifiable.

In a similar vein, although Gatsby is a surname as well and therefore could refer to a woman, the character of Jay Gatsby is also sufficiently iconic to make the word "Gatsby" imply "man." The alternate spelling "Gatsbie" looked too much like a child's nickname, so the final text uses the alternate spelling "Gatsbee" instead.

Character naming is highly subjective and readers may dislike the choices of new names in the example swapped novels.

**Side characters**

There are vastly more named characters in books than main characters. *Pride and Prejudice* contains over fifty character names, and *The Great Gatsby* contains over eighty. Aggregating all the names via a page-by-page read through is not efficient, so the software includes a function that automatically extracts all proper nouns into a table along with example sentences in which they appear, to facilitate identification of all side characters.

**Data availability statement**

As a demonstration of the Matriarchy Classics concept and method, *Gender-Swapped Pride and Prejudice* and *Gender-Swapped Great Gatsby* are completed and published as ebooks compatible with Kindle and other e-readers, available for free at www.matriarchyclassics.com. You are also welcome to directly email the corresponding author to obtain free copies of the ebooks.

**Collaboration**

The Matriarchy Classics method relies on extensive curated resources and implementation details refined over six years. The code is currently private to preserve the quality and integrity of the edition series. The proposed computational approach shortens the gender-swap time from weeks of manual labor to minutes, but several configuration steps remain manual, including creating character name gender swap mappings and reading through the final text. The author welcomes scholarly collaborations to create additional gender-swapped novels. If you would like to join the ongoing *Matriarchy Classics* project as a credited editor on future gender-swapped public domain novels, please email the author.

# Results

## Quantitative results

In an LLM-based method, it would be possible for any word of the text to arbitrarily change to any other word, due to the stochastic nature of the models. Thus, LLMs could produce both false positives (a genderless word is incorrectly changed to something else) and false negatives (failing to swap a gendered word). In the proposed Matriarchy Classics method, false positives are impossible, because the defined rules do not allow genderless words to change. Thus, only false negatives are possible.

*Pride and Prejudice* had a higher fraction of gendered words (14,257 of 122,264 = 11.7%) than *Gatsby* (4,152 of 48,661 = 8.5%). Of the 7,104 sentences in *Pride and Prejudice,* zero had automatically-detected pronoun swap errors, and ~15 sentences were manually edited during final human review which included reading multiple randomly-selected lengthy excerpts. Of the 3,407 sentences in *Gatsby,* 4 had automatically-detected pronoun swap errors, and ~38 sentences were manually edited during final human review, which included a beginning-to-end read of the full text. Of note, these manual edits included not only fixes of false negatives but also edits adjusting for cultural subtleties (e.g., cheekbones could be interpreted as gendered specifically in the context of shaving—discussed further in the next section). Overall, the false positive rate of the proposed Matriarchy Classics method is 0, and the sentence-level false negative rate can be estimated as (changed sentences) / (total sentences), or 15 / 7,104 = 0.0021 for *Pride and Prejudice* and 42 / 4,152 = 0.01 for *Gatsby.*

The proposed method yields a time savings of 48 to 122 hours per book, or 1.2 to 3 weeks of full-time labor per book (Table 2).

**Table 2. Human time savings from the proposed Matriarchy Classics method**
Gender-inverting the text by hand is cognitively taxing with a high error rate. The table estimates the time cost of a manual gender swap using an assumed rate of 1,000 words per hour, and book lengths of 48,661 words (*Gatsby*) and 122,264 words (*Pride and Prejudice*). The human time requirement for gender-swapping the book with the Matriarchy Classics method is the time it takes to type a single command on the command line (seconds) and so this is reported as "0 hours". The actual background runtime of the computational method is <5 minutes for the rule-based parts and a few hours for the pronoun-specific LLM query due to rate limits on the LLM API.

| Step | Manual gender swap | Matriarchy Classics method |
|---|---|---|
| Define character name mapping | 4 hours | |
| Gender-swap the book | 48 to 122 hours | 0 hours (automated) |
| Final proofread | 6 hours | |

## Manual edits analysis

The quantitative results section naturally raises the question, “What kind of manual edits were required?” This section explores some interesting examples from *Gatsby*.

In the original, Myrtle purchases “a small flask of perfume.” The word “perfume” was manually changed to “cologne” so that Merton purchases “a small flask of cologne.” This is an example of a straightforward false negative due to the word “perfume” not being present in the gender-swap mapping (it has since been added).

In the original, a man had just shaved and “there was a white spot of lather on his cheekbone […] I wiped from his cheek the spot of dried lather.” Typically the word “cheekbone” is not inherently gendered, but it could be interpreted as gendered in the context of shaving, because women do not commonly shave their faces. Because “cheekbone” and “cheek” can be considered contextually gendered here, they were manually edited to “ankle” implying leg-shaving instead of face-shaving: “there was a white spot of lather on her ankle […] I wiped from her ankle the spot of dried lather.” (This is unfortunately an anachronism because in the 1920s women typically concealed their legs with stockings and leg-shaving was not yet widespread, but in the spirit of the project all words with gendered implications must be swapped.)

In the original, “A celebrated tenor had sung in Italian, and a notorious contralto had sung in jazz.” Tenor was manually changed to alto, and contralto to countertenor.

One pronoun error represented a correct gender swap but an incorrect pronoun selection. The software changed “Daisy put her arm through his abruptly” to “David put his arm through her abruptly” which reads as horror-level violence. This was manually fixed to “David put his arm through *hers* abruptly.”

This line of dialogue from the original presents a dialectical challenge: “She ran out ina road. Son-of-a-bitch didn’t even stopus car.” “Stopus” is a dialectical rendering of “stop his” and in a gender swap ought to become “stop her” or, dialectically, something like “stopper.” The final sentence includes this manual edit to produce, “He ran out ina road. Daughter-of-a-bastard didn’t even stopper car.”

## Qualitative results

The best way to appreciate the qualitative result of this project, including its structural effects, character effects, and line-level effects, is to read the gender-swapped novels, which are available for free at www.matriarchyclassics.com. However, because reading an entire novel takes a few hours, this section provides a shorter illustration of some of the qualitative effects of gender inversion.

### Character effects

Inverting only gendered words in classic literature produces stories set in matriarchal societies, where women hold most positions of power and occupy narrative roles historically assigned to men. This section provides character descriptions for the gender-swapped novels, excerpting SparkNotes character descriptions and inverting only their gendered words[17,18].

*Gender-Swapped Pride and Prejudice* Partial Character List:

- Theodore Bennet: The novel's protagonist. The second son of Mrs. and Mr. Bennet, Theodore is the most intelligent and sensible of the five Bennet brothers.
- Wilhelmina Daphne: A wealthy lady, the mistress of Pemberley, and the niece of Lord Christopher de Bourgh. Though Daphne is intelligent and honest, her excess of pride causes her to look down on her social inferiors.
- Jonn Bennet: The eldest and most handsome Bennet brother. John is more reserved and gentler than Theodore.
- Caroline Bingley: Daphne's considerably wealthy best friend. She is a genial, well-intentioned lady, whose easygoing nature contrasts with Daphne's initially discourteous demeanor.
- Mrs. Bennet: The matriarch of the Bennet family, a lady of modest income with five unmarried sons. Mrs. Bennet has a sarcastic, cynical sense of humor that she uses to purposefully irritate her husband.
- Mr. Bennet: Mrs. Bennet's husband, a foolish, noisy man whose only goal in life is to see his sons married. Because of his low breeding and often unbecoming behavior, Mr. Bennet often repels the very suitors whom he tries to attract for his sons.

*Gender-Swapped Great Gatsby* Partial Character List:

- Nicole Carraway: The novel's narrator, Nicole is a young woman from Minnesota who, after being educated at Yale and fighting in World War I, goes to New York City to learn the bond business. Nicole quickly befriends her next-door neighbor, the mysterious Jane Gatsbee.
- Jane Gatsbee: The title character and protagonist of the novel, Gatsbee is a fabulously wealthy young woman living in a Gothic mansion in West Egg. She is famous for the lavish parties she throws every Saturday night.
- David Buchanan: Nicole's cousin, and the man Gatsbee loves. Now a handsome socialite, David lives with Thelma across from Gatsbee in the fashionable East Egg district of Long Island. He is sardonic and somewhat cynical, and behaves superficially to mask his pain at his wife's constant infidelity.
- Thelma Buchanan: David's immensely wealthy wife, once a member of Nicole's social club at Yale. Powerfully built and hailing from a socially solid old family, Thelma is an arrogant, hypocritical bully. Her social attitudes are laced with racism and sexism.

**Line-level effects**

Supplementary Tables S1 and S2 include examples quotes from the gender-swapped novels, such as Mrs. Bennet's criticism of two of her sons as "silly and ignorant like other boys" and Lydian daydreaming of visiting Brighton where he imagines himself "seated beneath a tent, tenderly flirting with at least six officers at once." From *Gatsby*, Thelma Buchanan is described thus: "Two shining arrogant eyes had established dominance over her face and gave her the appearance of always leaning aggressively forward." David remarks of her newborn son, "I'm glad it's a boy. And I hope he'll be a fool—that's the best thing a boy can be in this world, a handsome little fool."

# Discussion

This paper presents a computational approach to gender-swap classic literature in a trustworthy manner, with a programmatically-enforced guarantee that only gendered words are inverted. The method makes it more efficient to produce gender-swapped novels, reducing the amount of human effort required from weeks to minutes. The books produced through this method are defamiliarized[19–21] versions of the originals that facilitate analysis of gender roles and gendered assumptions embedded in both fiction and society.

## Pregnant men and the physical body

The rules of this experiment dictate that all gendered words must be swapped, and that *only* gendered words may be swapped. However, there is no "male equivalent" of pregnancy or childbirth, and so it is impossible to swap these words into something "male" that is the same. Therefore, passages related to pregnancy and childbearing were left untouched except for the gender of the person who is pregnant. Thus, we end up with jarring lines like the following in *Gatsby:* "The next April David had his little boy" and "Wilma was so sick that she looked guilty, unforgivably guilty—as if she had just got some poor boy with child."

At first glance this might seem like a failure of the experiment. But on further consideration, this is actually an interesting result—because it happens at all, and because it ends up mattering very little for the overall reading experience.

The fact that it's impossible to "gender-swap away" the notion of pregnant men connects with the thesis that physical reproductive burdens shape power dynamics between the sexes. In the real world, bodily challenges of pregnancy, childbirth, and lactation, placed upon women, are often considered a necessary (but not sufficient) condition for the development of patriarchy[22].

It is also noteworthy how little the notion of "pregnant men" matters for these particular books. From a storytelling and literary perspective, the novels function perfectly well after complete gender inversion. The books do not "break" or "fall apart" merely because women have power, money, authority, and control.

The project places women in the social role of men, and vice versa. This is in a different vein than speculative fiction or essays that imagine how female physical experiences in a patriarchal society would be perceived differently by society if they happened to males instead. Steinem's essay *If Men Could Menstruate* sits at the intersection of bodily processes and gender inversions[23]. It imagines a society where men menstruate, and menstruation is therefore enviable, boast-worthy, and publicly flaunted as a sign of power.

## Inheritance of money and property

In the gender-swapped classics, money and property are passed down through the female line. In *Gender-Swapped Great Gatsby* Nicole remarks, “We have a tradition that we’re descended from the Duchesses of Buccleuch, but the actual founder of my line was my grandmother’s sister, who came here in fifty-one, sent a substitute to the Civil War, and started the wholesale hardware business that my mother carries on today.” In these fictional matriarchal societies, gendered inheritance enhances female power[24].

## Limitations

This project does not currently incorporate other pronouns or genders beyond she/he and woman/man[25]. Because gendered terms were inverted, the romantic relationships in the swapped novels remain heterosexual. Future work could investigate swaps in which all characters become male or all characters become female, or characters are given another gender identity.

The current project does not focus on race or ethnicity. Hofstadter’s 1985 essay *A Person Paper on Purity in Language* coined new race-infused words like “mailwhite” (instead of mailman) and “whiteslaughter” (instead of manslaughter) to satirize those who oppose gender-inclusive language[26]. In English, race is not encoded in language in the same way that gender is encoded in language, so a “race swapped” text would either involve very few changes to the text (if only words directly related to race were changed), or extensive changes (if Hofstadter’s approach of inventing new words was employed instead).

The current project is English-specific, and does not address gender-swapping texts in additional languages.

## Future directions

A comprehensive analysis of the gender-swapped novels themselves is left to future work. The author intends to gender-swap additional public domain novels in the future, and welcomes collaborations towards this goal. It would be interesting to compare groups of gender-swapped novels across male and female authors, time periods, and genres.

Previously, the main limitation to gender-swapping long-form texts was time. With the proposed computational method, it now becomes possible to gender-swap any long-form text, with applications beyond public domain novels. This technique could be applied to public domain collections of short stories, or even biographies or history textbooks to explore a matriarchal alternate-history timeline. Feminist scholarly literature could itself be gender-swapped in this precise manner to explore how readers react to a feminist text versus the exact same text when it is instead about “masculinism.” Publishing houses could release gender-swapped versions of modern copyrighted novels. Web novels and fan fiction could also be swapped using the same method.

This paper does not include a formal reader response study. The one review posted on Goodreads for the Matriarchy Classics *Gender-Swapped Pride and Prejudice* notes, “it was just sometimes befuddling given my ingrained bias since I know the P&P world so well. Perhaps that was the point was trying to be made?" Future work could formally explore reader responses to gender-swapped novels, and could compare reader responses to the original novel versus the gender-swapped novel[12,13,27].

LLM reactions to content from gender-swapped novels could be used to probe gender biases encoded in AI.

**Other media**

With the power of generative AI in the future it may be possible to scalably apply this type of surgical-precision gender swap to all forms of media including photographs, paintings, TV shows, films, advertisements, and social media content[28]. All clothing, physical movements, and dialogue could remain identical with only the apparent gender of the characters changing. Since working on this project, the author has found it impossible to stop mentally gender-swapping scenarios across books, film, and social media, which has utterly ruined a large quantity of entertainment, albeit in a useful way.

# Conclusions

It is the author's hope that this computational framework and approach will be fertile ground for interrogating gender bias in classic literature and beyond.

# Supplementary Material

**Supplementary Table 1.** Before-and-after quotes demonstrating the gender inversion of *Pride and Prejudice.*

| Gender-Swapped Pride and Prejudice | Original Pride and Prejudice |
|---|---|
| *[Mrs. Bennet]* “They have none of them much to recommend them,” replied she: “they are all silly and ignorant like other boys; but Teddy has something more of quickness than his brothers.” | *[Mr. Bennet]* “They have none of them much to recommend them,” replied he: “they are all silly and ignorant like other girls; but Lizzy has something more of quickness than her sisters.” |
| Ms. Daphne soon drew the attention of the room by her fine, tall person, beautiful features, noble mien, and the report, which was in general circulation within five minutes after her entrance, of her having ten thousand a year. The ladies pronounced her to be a fine figure of a woman, the gentlemen declared she was much prettier than Ms. Beverley, and she was looked at with great admiration for about half the evening, till her manners gave a disgust which turned the tide of her popularity; for she was discovered to be proud, to be above her company, and above being pleased… | Mr. Darcy soon drew the attention of the room by his fine, tall person, handsome features, noble mien, and the report, which was in general circulation within five minutes after his entrance, of his having ten thousand a year. The gentlemen pronounced him to be a fine figure of a man, the ladies declared he was much handsomer than Mr. Bingley, and he was looked at with great admiration for about half the evening, till his manners gave a disgust which turned the tide of his popularity; for he was discovered to be proud, to be above his company, and above being pleased… |
| “I am not now to learn,” replied Ms. Colleen, with a formal wave of the hand, “that it is usual with young gentlemen to reject the addresses of the woman whom they secretly mean to accept, when she first applies for their favour; and that sometimes the refusal is repeated a second or even a third time. I am, therefore, by no means discouraged by what you have just said, and shall hope to lead you to the altar ere long.” | “I am not now to learn,” replied Mr. Collins, with a formal wave of the hand, “that it is usual with young ladies to reject the addresses of the man whom they secretly mean to accept, when he first applies for their favour; and that sometimes the refusal is repeated a second or even a third time. I am, therefore, by no means discouraged by what you have just said, and shall hope to lead you to the altar ere long.” |
| In Lydian’s imagination, a visit to Brighton comprised every possibility of earthly happiness. He saw, with the creative eye of fancy, the streets of that gay bathing-place covered with officers. He saw himself the object of attention to tens and to scores of them at present unknown. He saw all the glories of the camp: its tents stretched forth in | In Lydia’s imagination, a visit to Brighton comprised every possibility of earthly happiness. She saw, with the creative eye of fancy, the streets of that gay bathing-place covered with officers. She saw herself the object of attention to tens and to scores of them at present unknown. She saw all the glories of the camp: its tents stretched forth in |
| beauteous uniformity of lines, crowded with the young and the gay, and dazzling with scarlet; and, to complete the view, he saw himself seated beneath a tent, tenderly flirting with at least six officers at once. | beauteous uniformity of lines, crowded with the young and the gay, and dazzling with scarlet; and, to complete the view, she saw herself seated beneath a tent, tenderly flirting with at least six officers at once. |
| *[Ms. Daphne]* “If your feelings are still what they were last April, tell me so at once. My affections and wishes are unchanged; but one word from you will silence me on this subject for ever.”<br>Theodore, feeling all the more than common awkwardness and anxiety of her situation, now forced himself to speak; and immediately, though not very fluently, gave her to understand that his sentiments had undergone so material a change since the period to which she alluded, as to make him receive with gratitude and pleasure her present assurances. | *[Mr. Darcy]* “If your feelings are still what they were last April, tell me so at once. My affections and wishes are unchanged; but one word from you will silence me on this subject for ever.”<br>Elizabeth, feeling all the more than common awkwardness and anxiety of his situation, now forced herself to speak; and immediately, though not very fluently, gave him to understand that her sentiments had undergone so material a change since the period to which he alluded, as to make her receive with gratitude and pleasure his present assurances. |

**Supplementary Table 2.** Before-and-after quotes demonstrating the gender inversion of *The Great Gatsby.*

| **Gender-Swapped Great Gatsby** | **Original Great Gatsby** |
|---|---|
| …we have a tradition that we're descended from the Duchesses of Buccleuch, but the actual founder of my line was my grandmother's sister, who came here in fifty-one, sent a substitute to the Civil War, and started the wholesale hardware business that my mother carries on today. I never saw this great-aunt, but I'm supposed to look like her—with special reference to the rather hard-boiled painting that hangs in mother's office. | …we have a tradition that we're descended from the Dukes of Buccleuch, but the actual founder of my line was my grandfather's brother, who came here in fifty-one, sent a substitute to the Civil War, and started the wholesale hardware business that my father carries on today. I never saw this great-uncle, but I'm supposed to look like him—with special reference to the rather hard-boiled painting that hangs in father's office. |
| I had a dog—at least I had her for a few days until she ran away—and an old Dodge and a Finnish man, who made my bed and cooked breakfast and muttered Finnish wisdom to himself over the electric stove. | I had a dog—at least I had him for a few days until he ran away—and an old Dodge and a Finnish woman, who made my bed and cooked breakfast and muttered Finnish wisdom to herself over the electric stove. |
| *(of Thelma Buchanan)* Two shining arrogant eyes had established dominance over her face and gave her the appearance of always leaning aggressively forward. Not even the masculine swank of her riding clothes could hide the enormous power of that body—she seemed to fill those glistening boots until she strained the top lacing, and you could see a great pack of muscle shifting when her shoulder moved under her thin coat. It was a body capable of enormous leverage—a cruel body. | *(of Tom Buchanan)* Two shining arrogant eyes had established dominance over his face and gave him the appearance of always leaning aggressively forward. Not even the effeminate swank of his riding clothes could hide the enormous power of that body—he seemed to fill those glistening boots until he strained the top lacing, and you could see a great pack of muscle shifting when his shoulder moved under his thin coat. It was a body capable of enormous leverage—a cruel body. |
| 'All right,' I said, 'I'm glad it's a boy. And I hope he'll be a fool—that's the best thing a boy can be in this world, a handsome little fool.' | 'All right,' I said, 'I'm glad it's a girl. And I hope she'll be a fool—that's the best thing a girl can be in this world, a beautiful little fool.' |
| There was dancing now on the canvas in the garden; old women pushing young boys backward in eternal graceless circles… | There was dancing now on the canvas in the garden; old men pushing young girls backward in eternal graceless circles… |
| We were sitting at a table with a woman of about my age and a rowdy little boy, who gave way upon the slightest provocation to uncontrollable laughter. | We were sitting at a table with a man of about my age and a rowdy little girl, who gave way upon the slightest provocation to uncontrollable laughter. |
| He had drunk a quantity of champagne, and during the course of his song he had decided, ineptly, that everything was very, very sad—he was not only singing, he was weeping too. Whenever there was a pause in the song he filled it with gasping, broken sobs, and then took up the lyric again in a quavering bass. | She had drunk a quantity of champagne, and during the course of her song she had decided, ineptly, that everything was very, very sad—she was not only singing, she was weeping too. Whenever there was a pause in the song she filled it with gasping, broken sobs, and then took up the lyric again in a quavering soprano. |
| In spite of the husbands' agreement that such malevolence was beyond credibility, the dispute ended in a short struggle, and both husbands were lifted, kicking, into the night. | In spite of the wives' agreement that such malevolence was beyond credibility, the dispute ended in a short struggle, and both wives were lifted, kicking, into the night. |
| Some time toward midnight Thelma Buchanan and Mr. Wilma stood face to face discussing, in impassioned voices, whether Mr. Wilma had any right to mention David's name.<br>"David! David! David!" shouted Mr. Wilma. "I'll say it whenever I want to! David! Dav—"<br>Making a short deft movement, Thelma Buchanan broke his nose with her open hand. | Some time toward midnight Tom Buchanan and Mrs. Wilson stood face to face discussing, in impassioned voices, whether Mrs. Wilson had any right to mention Daisy's name.<br>"Daisy! Daisy! Daisy!" shouted Mrs. Wilson. "I'll say it whenever I want to! Daisy! Dai—"<br>Making a short deft movement, Tom Buchanan broke her nose with his open hand. |